%% file: PrefDPM-Provenance.tex
\documentclass[runningheads,a4paper]{llncs}

\usepackage{times}
\usepackage{helvet}
\usepackage{courier}

\usepackage{theorem}
\usepackage{graphicx}
\usepackage{latexsym}
\usepackage{amstext}
\usepackage{epsfig}
\usepackage{amsmath,amssymb}
\usepackage{amsfonts}

\usepackage{subfigure}

\usepackage{multirow}

\usepackage{url}

\usepackage{color}

\input{definitions}

\input{preamble}

\title{Top-k Query Answering in \dpm\ Ontologies\\ under
Subjective Reports (Technical Report)}

\titlerunning{Top-k Query Answering under
Subjective Reports (Technical Report)}

\author{Thomas Lukasiewicz\inst{1} \and
Maria Vanina Martinez\inst{1} \and
Cristian Molinaro\inst{2} \and \\
Livia Predoiu\inst{1} \and
Gerardo I.\ Simari\inst{1} 
}

\authorrunning{T.~Lukasiewicz, M.V.~Martinez, C.\ Molinaro, L.\ Predoiu, and G.I.~Simari}

\institute{Department of Computer Science, University of Oxford, UK\\
\email{\{thomas.lukasiewicz,vanina.martinez,livia.predoiu, gerardo.simari\}@cs.ox.ac.uk}
\and
DIMES, Universit\`a della Calabria, Italy \\
\email{cmolinaro@dimes.unical.it}
}

\begin{document}

\maketitle

\begin{abstract}
The use of preferences in query answering, both in traditional data\-bases and in ontology-based data access, has recently received
much attention, due to its many real-world applications.
In this paper, we tackle the problem of top-$k$ query answering in \dpm\ ontologies subject to the querying user's preferences
and a collection of (subjective) reports of other users. Here,
each report consists of scores for a list of features,
its author's preferences among the features, as well as other information. Theses pieces of information of every report are then combined, along with the querying user's preferences and his/her trust into each report, to rank the query results. We present two alternative such rankings, along with algorithms for top-$k$ (atomic) query answering under these rankings. We also show that, under suitable assumptions, these algorithms run in polynomial time in the data complexity. We finally present more general reports, which~are associated with sets of atoms rather than single atoms.
\end{abstract}

\noindent



\input{intro-rw}


\input{preliminaries}


\input{reports}


\input{queryans}


\input{gen-reports}


\input{summary}


\smallskip
\noindent
{\bf Acknowledgments.}
This work was supported by the UK EPSRC grant EP/J008346/1 \hbox{(``PrOQAW'')}, an EU (FP7/2007-2013) Marie-Curie Intra-European Fellowship, the ERC grant 246858 (``DIADEM''), and a Yahoo! Research Fellowship.


\bibliographystyle{splncs}
\bibliography{PrefDatalog,ProbPref}


\end{document}

%% file: definitions.tex
\newcommand{\PU}{\succ_{P_U}}
\def\<{\langle}
\def\>{\rangle}
\def\sq{simple}
\def\mg{\sqsubseteq}

\newcommand{\PredOnt}{\R}

\newcommand{\PredPref}{\R_{\textit{Pref}}}

\newcommand{\ConsPref}{\Delta_{\textit{Pref}}}

\newcommand{\VarsPref}{\V_{\textit{Pref}}}

\newcommand{\HerbOnt}{\mathcal{H}}
\newcommand{\HerbPref}{\mathcal{H}_{\textit{Pref}}}

\newcommand{\smallbsq}{\hfill{\tiny $\blacksquare$}}

\newcommand{\dpm}{Datalog{+}/{--}}

\newcommand{\reprankbasic}[0]{{\sf RepRank-Basic}}
\newcommand{\reprankhist}[0]{{\sf RepRank-Hist}}

\newcommand{\summarizeReports}[0]{{\sf SummarizeReports}}

\newcommand{\loc}{\textit{loc}}
\newcommand{\cl}{\textit{cl}}
\newcommand{\pri}{\textit{pri}}

\newcommand{\br}{\textit{br}}
\newcommand{\net}{\textit{net}}

\newcommand{\kf}{\textit{kfac}}

\newcommand{\nop}[1]{}

 \newcommand{\F}{\mathcal{F}}

 \newcommand{\R}{\mathcal{R}}
 
 \newcommand{\V}{\mathcal{V}}

\newcommand{\KB}{\mathit{K\!B}}
\newcommand{\eqs}{{\,{=}\,}}
\newcommand{\ins}{{\,{\in}\,}}

\newcommand{\ges}{\,{\ge}\,}

\newcommand{\ras}{\,{\ra}\,}

\newcommand{\cups}{\,{\cup}\,}

\newcommand{\subseteqs}{\,{\subseteq}\,}

\newcommand{\cR}{{\cal R}}

\renewcommand{\ge}{\geqslant}

\renewcommand{\geq}{\geqslant}
\renewcommand{\leq}{\leqslant}

\newcommand{\ra}{\rightarrow}

\newcommand{\dd}[2]{#1_1,\ldots,#1_{#2}}

\newcommand{\vett}[1]{\mathbf{#1}}

\newcommand{\sol}[2]{\mathit{mods}(#1,#2)}
\newcommand{\dom}{\Delta}
\newcommand{\freshdom}{\Delta_N}
\newcommand{\variables}{\mathcal{V}}

\newcommand{\dep}{\Sigma}

\newcommand{\isa}[1]{\mathit{ISA}}

\newcommand{\head}[1]{\mathit{head}(#1)}
\newcommand{\body}[1]{\mathit{body}(#1)}

\newcommand{\ans}[3]{\mathit{ans}(#1,#2,#3)}

\newcommand{\atom}[1]{#1}

\newcommand{\chase}[2]{\mathit{chase}(#1,#2)}

\newcounter{cefalo}

\newcounter{cefalocont}

\def\qed{\hfill{\qedboxempty}
  \ifdim\lastskip<\medskipamount \removelastskip\penalty55\medskip\fi}

\def\qedboxempty{\vbox{\hrule\hbox{\vrule\kern3pt
                 \vbox{\kern3pt\kern3pt}\kern3pt\vrule}\hrule}}

\def\qedfull{\hfill{\qedboxfull}
  \ifdim\lastskip<\medskipamount \removelastskip\penalty55\medskip\fi}

\def\qedboxfull{\vrule height 4pt width 4pt depth 0pt}

%% file: preamble.tex
\long\def\symbolfootnote[#1]#2{\begingroup%
\def\thefootnote{\fnsymbol{footnote}}\footnote[#1]{#2}\endgroup}

%% file: intro-rw.tex
\section{Introduction}
\label{sec:intro-rw}

The use of preferences in query answering, both in traditional data\-bases and in ontology-based data access, has recently received
much attention due to its many real-world applications. In particular, in recent times, there has been a
huge change in the way data is created and consumed, and users have largely moved to the
Social Web, a system of platforms used to socially interact by sharing data and collaborating on tasks.

In this paper, we tackle the problem of preference-based query answering in Data\-log+/-- ontologies assuming that the user
must rely on subjective reports to get a complete picture and make a decision.
This kind of situation arises all the time on the Web; for instance, when searching for a hotel, users provide some basic information and receive a list of answers to choose from, each associated with a set of subjective reports (often called reviews) written by other users to tell everyone about their experience.
The main problem with this setup, however, is that users are often overwhelmed and frustrated, because they cannot decide which reviews to focus on and which ones to ignore, since it is likely that, for instance, a very negative (or positive) review may have been produced on the basis of a feature that is completely irrelevant to the querying user.

We study a formalization of this process and its incorporation into
preference-based query answering in \dpm\ ontologies, proposing the use of trust and relevance measures to select the best reports to focus on, given the user's initial preferences, as well as novel ranking algorithms to obtain a user-tailored answer.
The main contributions of this paper can be briefly summarized as follows.
\begin{itemize}
\item We present an approach to preference-based top-$k$ query answering in Datalog+/-- ontologies, given a collection of subjective reports. Here, each report contains scores for a list of features,
its author's preferences among the features, as well as additional information. Theses pieces of information of every report are then aggregated, along with the querying user's trust into each report, to a ranking of the~query results relative to the preferences of the querying user.
\item We present a basic approach to ranking the query results, where each atom is associated with the average of the scores of all reports, and every report is ranked with the average of the scores of each feature, weighted by the report's trust values and the relevance of the feature and of the report for the querying user.
\item We then present an alternative approach to ranking the query results, where we first select the most relevant reports for the querying user, adjust the scores by the trust measure, and
compute a single score for each atom by combining the scores computed in the previous step, weighted by the relevance of the features.
\item We present algorithms for preference-based top-$k$ (atomic) query answering in Data\-log+/-- ontologies under both rankings. We also prove that, under suitable assumptions, the two algorithms run in polynomial time in the data complexity.
\item Finally, we also propose and discuss a more general form of reports, which are associated with sets of atoms rather than single atoms.
\end{itemize}

The rest of this paper is organized as follows. In Section~\ref{sec:prelims}, we provide some preliminaries on Data\-log+/-- and the used preference models. Section~\ref{sec:reports}
then defines subjective reports, along with their trust measures and their relevance. In Section~\ref{sec:QA-basic}, we introduce the two rankings of query results, along with top-$k$ query answering algorithms under these rankings and data tractability results.
Section~\ref{sec:general} then presents more general subjective reports. In Section~\ref{sec:rel}, we discuss related work. Finally, the concluding Section~\ref{sec:conc} summarizes the main results of this paper and gives an outlook on future research.


%% file: preliminaries.tex
\section{Preliminaries}
\label{sec:prelims}



First, we briefly recall some basics on
\dpm~\cite{JWSDatalog:2011}, namely, on relational data\-bases and  (Boolean) conjunctive queries ((B)CQs) (along with 
tuple- and equality-gene\-r\-ating dependencies (TGDs and EGDs, respectively) and 
negative constraints), the chase procedure, and ontologies in \dpm.
We also define the used preference models. 

\medskip
\noindent
{\bf Databases and Queries}.
We assume {(i)} an
infinite universe of \emph{(data) constants} $\dom$ (which constitute the ``normal''
domain of a database), {(ii)}~an infinite set of
\textit{(labeled) nulls} $\freshdom$ (used as ``fresh'' Skolem terms,
which are placeholders for unknown values, and can thus be seen as variables),
and {(iii)} an infinite set of variables~$\variables$
(used in queries, dependencies, and constraints). Different constants represent
different values (\emph{unique name assumption}),
while different nulls may represent the same value.
We assume a lexicographic order on $\dom\cup\freshdom$, with every symbol in~$\freshdom$ following all symbols in $\dom$.
We denote by~$\vett{X}$ sequences of variables $\dd{X}{k}$ with~$k\ges 0$.
We assume a {\em relational schema} $\cR$, which is a finite set of {\em predicate symbols} (or simply {\em predicates}).
A~{\em term}~$t$ is a constant, null, or variable.
An {\em atomic formula} (or {\em atom}) $\atom{a}$ has the form~$P(t_1,...,t_n)$, where~$P$ is an $n$-ary predicate, and~$t_1,...,t_n$ are terms.
We say that $\atom{a}$ is \emph{ground} iff every~$t_i$ belongs to $\dom$. 

A {\em database (instance)} $D$ for a relational schema~$\R$ is a (possibly infinite)
set of atoms with predicates from $\R$ and arguments from~$\dom$.
A {\em conjunctive query (CQ)} over~$\R$ has the form $Q(\vett{X})=\exists\vett{Y}\,\Phi(\vett{X},\vett{Y})$,
where $\Phi(\vett{X},\vett{Y})$ is a conjunction of atoms
(possibly equalities, but not inequalities) with the variables $\vett{X}$ and~$\vett{Y}$, and possibly constants, but no nulls.
A CQ is \emph{atomic} iff $\Phi(\vett{X},\vett{Y})$ is a single atom and $\vett{Y}\eqs \emptyset$ (i.e., there are no existentially quantified variables).
A {\em Bool\-ean CQ} (BCQ) over $\R$ is a CQ of the form $Q()$, i.e., all variables are existentially quantified, often written as the set of all its atoms without quantifiers, 
when there is no danger of confusion. 
Answers to CQs and BCQs are defined via {\em homomorphisms}, which are
mappings $\mu\colon \dom \cup
\freshdom\cup\variables\rightarrow\dom \cup \freshdom\cup\variables$ such that
(i) $c \ins \dom$ implies~$\mu(c)\eqs c$,
(ii)~$c \ins \freshdom$ implies~$\mu(c) \ins \dom\cups\freshdom$,
and (iii)~$\mu$~is naturally extended to atoms, sets of atoms,
and conjunctions of atoms.
The set of all {\em answers} to a CQ $Q(\vett{X})\,{=}\,\exists\vett{Y}\,\Phi(\vett{X},\vett{Y})$ over $D$, denoted~$Q(D)$, is the set of all tuples $\atom{t}$ over $\dom$ for which there exists a homomorphism $\mu\colon \vett{X}\cups \vett{Y}\ras \dom \cup \freshdom$ such that $\mu(\Phi(\vett{X},\vett{Y}))\subseteqs D$ and $\mu(\vett{X})\eqs \atom{t}$. 
The {\em answer}~to a BCQ~$Q()$ over a database~$D$ is~{\em Yes}, denoted~$D\,{\models}\, Q$, iff $Q(D)\,{\neq}\,\emptyset$. 

Given a relational schema~$\R$, a \emph{tuple-gene\-r\-ating dependency} (TGD)
$\sigma$ is a first-order formula of the form $\forall \vett{X}\forall \vett{Y}\,\Phi(\vett{X},\vett{Y}) \ra
\exists{\vett{Z}} \,\Psi(\vett{X},\vett{Z})$, where $\Phi(\vett{X},\vett{Y})$
and $\Psi(\vett{X},$ $\vett{Z})$ are conjunctions of atoms over~$\R$
(without nulls), called
the \emph{body} and the \emph{head} of $\sigma$, denoted $\body{\sigma}$ and
$\head{\sigma}$, respectively.
Such~$\sigma$ is satisfied in a database~$D$ for~$\R$ iff, whenever there exists a homomorphism $h$ that maps the atoms of
$\Phi(\vett{X},\vett{Y})$ to atoms of $D$, there exists
an extension~$h'$ of
$h$ that maps the atoms of~$\Psi(\vett{X},\vett{Z})$ to atoms of $D$.
All sets of TGDs are finite here. Since TGDs can be reduced to TGDs with only single atoms in their heads, in the sequel,
every TGD has w.l.o.g.\ a single atom in its head.
A TGD $\sigma$ is {\em guarded} iff it contains an atom in its body
that contains all universally quantified variables of $\sigma$.
The leftmost such atom is the~{\em guard atom} (or {\em guard}) of $\sigma$.
A~TGD $\sigma$ is {\em linear} iff it contains only a single atom in its body.
As set of TGDs is guarded (resp., linear) iff all its TGDs are  guarded (resp., linear).

{\em Query answering} under TGDs, i.e.,
the evaluation of CQs and BCQs on data\-ba\-ses under a set of TGDs is
defined as follows. For a database $D$ for $\R$, and a set of TGDs $\dep$ on $\R$, the set of \emph{models} of $D$ and $\dep$,
denoted $\sol{D}{\dep}$, is the set of all (possibly infinite) databases $B$ such that (i) $D\subseteqs B$ and
(ii) every $\sigma\,{\in}\,\dep$ is satisfied in $B$.
The set of \emph{answers} for a CQ $Q$ to $D$ and $\dep$, denoted 
$\ans{Q}{D}{\dep}$ (or, for $\KB \eqs (D,\Sigma)$, $\textit{ans}(Q,\KB)$), is the set of all tuples
$t$ such that $t \in Q(B)$ for all $B \,{\in}\, \sol{D}{\dep}$.
The {\em answer} for a BCQ~$Q$ to $D$ and $\dep$ is {\em Yes}, denoted~$D\cup \Sigma\,{\models}\, Q$, iff
$\ans{Q}{D}{\dep}\,{\neq}\,\emptyset$.
Note that query answering under general TGDs is undecidable~\cite{BeVa81},
even when the schema and TGDs are fixed~\cite{CaGK08}.
Decidability and tractability in the data complexity of query answering for the guarded case follows from a bounded tree-width property. 

A {\em negative constraint} (or simply {\em constraint}) $\gamma$ is a
first-order formula of the form $\forall \vett{X} \Phi(\vett{X}) \,{\rightarrow}\, \bot$, where $\Phi(\vett{X})$
(called the {\em body} of $\gamma$) is a conjunction of atoms over $\R$ (without nulls).
Under the standard semantics of query answering of BCQs in \dpm\ with TGDs, adding negative constraints is computationally easy,
as for each constraint $\forall\vett{X} \Phi(\vett{X}) \,{\rightarrow}\, \bot$, we only have to check that the BCQ
$\exists\vett{X}\,\Phi(\vett{X})$ 
evaluates to false in~$D$ under $\dep$; if one of these checks fails, then
the answer to the original BCQ $Q$ is true, otherwise
the constraints can simply be ignored when answering the
BCQ~$Q$. 

An \emph{equality-generating dependency} (EGD)~$\sigma$ is a first-order formula of the form $\forall\vett{X}\,\Phi(\vett{X})$ $\ras X_i\eqs X_j$, where
$\Phi(\vett{X})$, called the \emph{body} of $\sigma$ and denoted $\body{\sigma}$, is a
conjunction of atoms over $\R$ (without nulls), and $X_i$ and~$X_j$ are variables from $\vett{X}$. Such~$\sigma$ is satisfied in a
database~$D$ for $\R$ iff, whenever there is a homomorphism $h$
such that $h(\Phi(\vett{X},\vett{Y}))\subseteqs D$, it holds that $h(X_i)\eqs h(X_j)$.
Adding EGDs over databases with TGDs along with negative constraints
does not increase the complexity of BCQ query answering as long as they are
{\em non-conflicting} \cite{JWSDatalog:2011}. Intuitively, this ensures that, if the chase (see below) fails
(due to strong violations of EGDs), then it already fails on the database, and if it does not fail, then whenever ``new'' atoms
are created in the chase by the application of
the EGD chase rule, atoms that are logically equivalent to the new ones are
guaranteed to be generated also in the absence of the EGDs, guaranteeing that EGDs do not influence
the chase with respect to query answering.

We usually omit the universal quantifiers in TGDs, negative constraints, and EGDs,
and we implicitly assume that all sets of dependencies and/or constraints are finite.

\medskip
\noindent
{\bf The Chase.}
The \emph{chase} was first introduced to enable checking implication of dependencies,
and later also for checking query containment.
By ``chase'', we refer both to the chase procedure and to its output.
The TGD chase works on a database via so-called TGD \emph{chase rules} (see \cite{JWSDatalog:2011} for an
extended chase with also EGD chase rules).

{\em TGD Chase Rule.} Let $D$ be a data\-base, and $\sigma$ a TGD of the form $\Phi(\vett{X},\vett{Y})
\ra \exists{\vett{Z}}\,  \Psi(\vett{X},$ $\vett{Z})$.  Then, $\sigma$ is \emph{applicable} to~$D$ iff
there exists a homomorphism $h$ that maps the atoms of $\Phi(\vett{X},\vett{Y})$ to
atoms of~$D$. Let $\sigma$ be applicable to $D$, and $h_1$ be a homomorphism that
extends $h$ as follows: for each $X_i\in \vett{X}$, $h_1(X_i)=h(X_i)$; for each
$Z_j\in \vett{Z}$, $h_1(Z_j)=z_j$, where $z_j$ is a ``fresh'' null, i.e.,
$z_j\in \freshdom$, $z_j$ does not occur in $D$, and $z_j$ lexicographically follows all
other nulls already introduced.
The {\em application of} $\sigma$ on $D$ adds to~$D$ the atom $h_1(\Psi(\vett{X},\vett{Z}))$ if not already
in~$D$.

The chase algorithm for a database $D$ and a set of TGDs $\Sigma$ consists of an exhaustive application of the
TGD chase rule in a breadth-first (level-sat\-urating) fashion,
which outputs a (possibly infinite) chase for $D$ and~$\Sigma$.
Formally, the {\em chase of level up to~$0$} of $D$ relative to~$\Sigma$, de\-no\-ted~$\textit{chase}^0(D,\Sigma)$,
is defined as $D$, assigning to every atom in $D$ the \emph{(derivation) level} $0$.
For every $k\ges 1$, the~{\em chase of level up to~$k$} of~$D$
relative to~$\Sigma$, de\-no\-ted~$\textit{chase}^k(D,\Sigma)$, is constructed as follows:
let $\dd{I}{n}$ be all possible images of bodies of TGDs in $\dep$ relative to some
homomorphism such that (i)~$\dd{I}{n}\subseteqs \textit{chase}^{k-1}(D,\Sigma)$
and (ii)~the highest level of an atom in every $I_i$ is $k-1$;
then, perform every corresponding TGD application on $\textit{chase}^{k-1}(D,\Sigma)$,
choosing the applied TGDs and homomorphisms in a (fixed) linear and lexicographic order, respectively,
and assigning to every new atom the \emph{(derivation) level} $k$.
The {\em chase} of $D$ relative to  $\Sigma$, denoted $\chase{D}{\dep}$, is defined as the limit
of $\textit{chase}^k(D,\Sigma)$ for~$k\,{\rightarrow}\,\infty$.

The (possibly infinite) chase relative to TGDs is a
\emph{universal mo\-del}, i.e., there exists a homomorphism from $\chase{D}{\dep}$ onto every
$B \,{\in}\, \sol{D}{\dep}$~\cite{JWSDatalog:2011}.
This implies that BCQs $Q$ over $D$ and $\Sigma$ can be evaluated on the chase
for~$D$ and $\Sigma$, i.e., $D\cups \Sigma \models Q$ is equivalent to $\chase{D}{\dep}\models Q$.
For guarded TGDs~$\Sigma$, such BCQs $Q$ can be evaluated on an initial fragment of
$\chase{D}{\dep}$ of constant depth~$k \cdot |Q|$, which
is possible in polynomial time in the data complexity.

\medskip
\noindent
{\bf \dpm\ Ontologies.}
A {\em \dpm\ ontology}
$\KB \eqs (D, \Sigma )$, where $\Sigma \eqs \Sigma_T \cup \Sigma_E \cup \Sigma_{\textit{NC}}$,
consists of a database $D$, a set of TGDs $\Sigma_T$, a~set of
non-conflicting EGDs $\Sigma_E$, and a set of negative constraints~$\Sigma_{\textit{NC}}$. 
We say that $\KB$ is {\em guarded} (resp., {\em linear}) iff
$\Sigma_T$ is guarded (resp., linear).
The following example illustrates a simple \dpm\ ontology, which
is used in the sequel as a running example.

\begin{example}
\label{ex:running}
{\rm
\label{ex:dpm}
Consider the following simple ontology $\KB = ( D, \Sigma)$, where:
\[
\begin{array}{rcll}
\Sigma & =&\{\sigma_1: \textit{hotel}(H) \rightarrow \textit{accom}(H), \\
       & & \phantom{\{}\sigma_2: \textit{apartment}(A) \rightarrow \textit{accom}(A), \\
       & & \phantom{\{}\sigma_3: \textit{bb}(B) \rightarrow \textit{accom}(B), \\
       & & \phantom{\{}\sigma_4: \textit{apthotel}(A) \rightarrow \textit{hotel}(A), \\
       & & \phantom{\{}\sigma_5: \textit{hostel}(H) \rightarrow \exists B \, \textit{bed}(B,H), \\
       & & \phantom{\{}\sigma_6: \textit{hotel}(H) \rightarrow \exists R \, \textit{room}(R,H), \\
       & & \phantom{\{}\sigma_7: \textit{bb}(B) \rightarrow \exists R \, \textit{room}(R,B) \} \\
\end{array}\]
and $D \eqs \{
\textit{hotel}(h_1),$ $
\textit{hotel}(h_2),$ $ 
\textit{locatedIn}(h_1,\textit{oxford}),$ $\textit{locatedIn}(h_2,\textit{oxfordCenter}),$ \linebreak
$\textit{hostel}(hs_1),
\textit{bb}(bb_1),
\textit{apartment}(a_1),
\textit{apthotel}(a_2),
\textit{locatedIn}(a_2,\textit{oxfordCenter})
\}$.

\smallskip
This ontology models a very simple accommodation booking domain, which could be used
as the underlying model in an online system. Accommodations can be either hotels,
bed and breakfasts, hostels, apartments, or aparthotel.
The database $D$ provides some instances for each kind of accommodation, as well as some location facts. 
\smallbsq
}
\end{example}

%

\medskip
\noindent
{\bf Preference Models}.
We now briefly recall some basic concepts regarding the representation of preferences.
We assume the following sets, giving rise to the logical language used for this purpose:
$\ConsPref\subseteq \Delta$ is a finite set of constants,
$\PredPref\subseteq \PredOnt$ is finite set of predicates,
and $\VarsPref\subseteq \mathcal{V}$ is an infinite sets of variables.
These sets give rise to a corresponding {\em Herbrand base} consisting
of all possible ground atoms that can be formed, which
we denote with $\HerbPref$, while $\HerbOnt$ is the Herbrand base for the ontology.
Clearly, we have $\HerbPref \subseteq \HerbOnt$, meaning that
preference relations are defined over a subset of the possible ground atoms.

A {\em preference relation} over set $S$ is any binary relation $\succ \; \subseteq S \times S$.
Here, we are interested in
strict partial orders (SPOs), which are irreflexive and transitive relations---we consider these to be
the minimal requirements for a useful preference relation.
One possible way of specifying such a relation is the preference
formula framework of~\cite{Chomicki:2003}.
We use $\textit{SPOs}(S)$ to denote the set of all possible strict partial orders over a set $S$.

Finally, the {\em rank} of an element in a preference relation $\succ$ is defined inductively as follows:
(i) $\textit{rank}(a,\succ) = 1$ iff there is no $b$ such that $b \succ a$; and
(ii) $\textit{rank}(a,\succ) = k + 1$ iff $\textit{rank}(a,\succ) = 1$ after eliminating from $\succ$ all elements of
rank at most $k$. 

%% file: reports.tex
\section{Subjective Reports}
\label{sec:reports}

Let $\KB$ be a \dpm\ ontology, $a = p(c_1,...,c_m)$ be a ground atom such that $\KB \models \atom{a}$,  and $\F = (f_1,..., f_n)$ be a tuple of {\em features} associated with the predicate $p$, each of which has  a domain $\textit{dom}(f_i) = [0,1] \cup \{-\}$.
We sometimes slightly abuse notation and use $\F$ to also denote the set of features $\{f_1,...,f_n\}$.

\begin{definition}{\rm
\label{def:report}
A {\em report} for a ground atom $\atom{a}$ is a triple $(E,\succ_P,I)$, where $E \in \textit{dom}(f_1) \times ... \times \textit{dom}(f_n)$,
$\succ_P$ is an SPO over the elements of $\F$, and $I$ is a set of pairs $(\textit{key}, \textit{value})$.
}\end{definition}

Intuitively, reports are evaluations of an entity of interest (atom $\atom{a}$) provided by observers.
In a report $(E,\succ_P,I)$, $E$ specifies a ``score'' for each feature, $\succ_P$ indicates the relative
importance of the features to the report's observer, and~$I$ (called \emph{information register})
contains general information about the report itself and who provided it.
Reports will be analyzed by a user, who has his own strict partial order, denoted $\PU$, over the set of features.
The following is a simple example involving hotel ratings.

\begin{figure}[t]
\centering
\includegraphics[width=0.8\columnwidth]{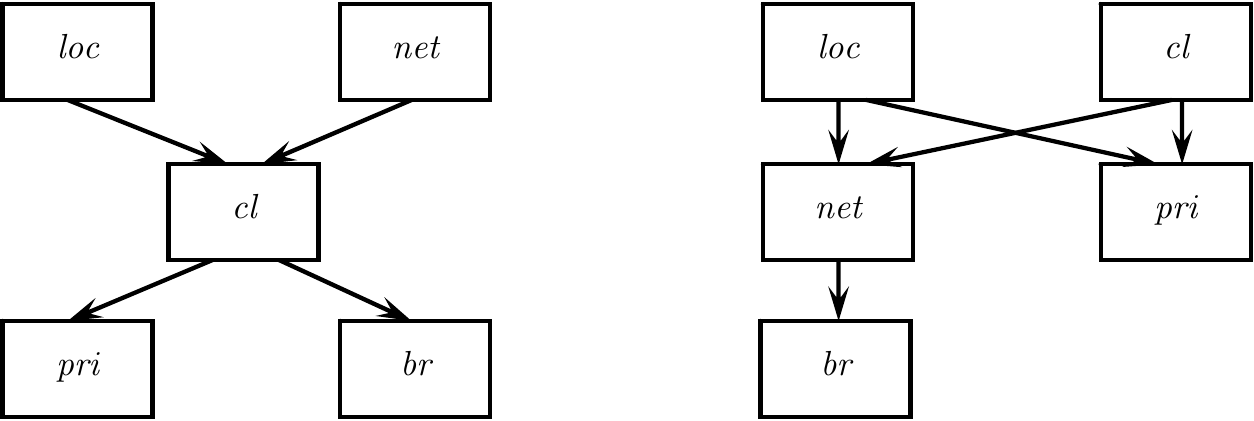}
\caption{Preference relation $\succ_{P_1}$ from Example~\ref{ex:reports} (left) and $\PU$, the user's preferences (right).}
\label{fig:reports}
\end{figure}

\begin{example}
\label{ex:reports}
Consider again the accommodation domain from Example~\ref{ex:running}, and
let the features for predicate $\textit{hotel}$ be
$\F = (\textit{location},$
$\textit{cleanliness},$
$\textit{price},$
$\textit{breakfast},$
$\textit{internet})$; in the following, we abbreviate these features as
{\em loc}, {\em cl}, {\em pri}, {\em br}, and {\em net}, respectively.

An example of a report for $\textit{hotel}(h_1)$ is
$r_1 \eqs (\langle 1,0,0.4,0.1,1\rangle, \succ_{P_1}, I_1)$, where
$\succ_{P_1}$ is given by the graph in Fig.~\ref{fig:reports} (left side); $\PU$ (the user's SPO) is shown in the same
figure (right side). Finally, let $I_1$ be a register with fields {\em age}, {\em nationality}, and
{\em type of traveler},
with data $I_1.\textit{age} = 34$, $I_1.\textit{nationality} = \textit{Italian}$, and
$I_1.\textit{type} = \textit{Business}$.
\smallbsq
\end{example}

The set of all reports available is denoted with $\textit{Reports}$.
In the following, we use $\textit{Reports}(\atom{a})$ to denote the set of all reports that are associated with a ground atom $\atom{a}$.
Given a tuple of features $\F$, we use $\textit{SPOs}(\F)$ to denote the set of all SPOs over $\F$.


\subsection{Trust Measures over Reports}
\label{sec:trust}

A user analyzing a set of reports may decide that certain opinions within a given report may be more trustworthy
than others. For instance, returning to our running example, the score given for the {\em location} feature of
$\textit{hotel}(h_1)$ might be considered more trustworthy than the ones given for {\em price} or
{\em breakfast}, e.g., because the report declared the former to be among the most preferred
features, while the latter are among the least preferred ones, cf.\ Figure~\ref{fig:reports} (left).
Another example could be a user that is generally untrustworthy of reports on feature {\em cleanliness}, because
he has learned that people are in general much more critical than he is when it comes to evaluating that aspect
of a hotel, or of reports on feature {\em price} by business travelers because they do not use their own money to  pay.
Formally, we have the following definition of trust measure.

\begin{definition}{\rm
\label{def:trustMeasure}
A {\em trust measure} is any function $\tau: \textit{Reports} \rightarrow [0,1]^n$.
}\end{definition}

Note that trust measures do not depend on the user's own preferences over $\F$ (in $\PU$); rather, for each report $(E,\succ_P,I)$, they give a measure of trust to each of the $n$ scores in $E$ depending on $P$ and $I$.
The following shows an example of a trust measure.

\begin{example}
\label{ex:trust}
Consider again our running example and suppose that the user defines a trust measure $\tau$, which assigns trust values to a report $r=(E,\succ_P,I)$ as follows:
\[
\tau(r) =
\left\{
	\begin{array}{ll}
		0.25 \cdot \big(2^{-(\textit{rank}(f_1,\succ_P)-1)}, ..., 2^{-(\textit{rank}(f_n,\succ_P)-1)}\big)  & \ \ \textit{if} \; I.\textit{nationality} \neq \textit{Italian}\textrm{;} \\
		\big(2^{-(\textit{rank}(f_1,\succ_P)-1)}, ..., 2^{-(\textit{rank}(f_n,\succ_P)-1)}\big) & \ \ \textrm{otherwise.}
	\end{array}
\right.
\]
For $r_1$ from Example~\ref{ex:reports} and the SPO in  Fig.~\ref{fig:reports} (left side), we get
$2^{-(\textit{rank}(loc,\succ_{P_1})-1)} = 2^{-(\textit{rank}(net,\succ_{P_1})-1)} = 1$,
$2^{-(\textit{rank}(cl,\succ_{P_1})-1)} = 0.5$, and $2^{-(\textit{rank}(pri,\succ_{P_1})-1)} = 0.25$.
\smallbsq
\end{example}


\subsection{Relevance of Reports}
\label{sec:relevance}

The other aspect of importance that a user must consider when analyzing reports is how {\em relevant} they are to his/her own preferences.
For instance, a report given by someone who has preferences that are completely opposite to those of the user should be considered less relevant than one given by someone whose preferences only differ in a
trivial aspect. This is inherently different from the trust measure described above, since trust is computed without
taking into account the preference relation given by the user issuing the query. Formally, we define relevance measures
as follows.

\begin{definition}{\rm
\label{def:relevanceMeasure}
A {\em relevance measure} is any function
$\rho: \textit{Reports} \times \textit{SPOs}(\F) \rightarrow [0,1]$.
}\end{definition}

Thus, a relevance measure takes as input a report $(E,\succ_P,I)$ and an SPO $\succ_{P'}$ and gives a measure of how relevant the report is relative to $\succ_{P'}$; this is determined on the basis of $\succ_P$ and $\succ_{P'}$, and can also take $I$ into account.

\begin{example}
\label{ex:relevance}
Consider again the running example, and suppose that the user assigns relevance to a report $r=(E,\succ_P,I)$ according to
the function $$\rho(r,\PU) = 2^{-\sum_{f_i \in \F} |\textit{rank}(f_i,\succ_P)-\textit{rank}(f_i,\PU)|}.$$
\noindent
From Fig.~\ref{fig:reports}, e.g., we have that
$\rho(r_1,\PU) = 2^{-1 * (0 + 1 + 1 + 0 + 1)} = 0.125$.

Alternatively, a relevance measure comparing the SPO $\succ_P$ of a report $(E,\succ_P,I)$ with the user's SPO $\PU$ (thus, in this case, information in $I$ is ignored by the relevance measure) might be defined as follows.
The relevance measure checks to what extent the two SPOs agree on the relative importance of the features in $\F$.
Formally, let $P_1$ and $P_2$ be SPOs over $\F$.
We define a measure of similarity of $P_1$ and $P_2$ as follows:
\[
sim(P_1,P_2)=\frac{\sum\limits_{1\leq i<j \leq n} sim(f_i,f_j,P_1,P_2)}{n(n-1)/2}\,,
\]
where
\[
sim(f_i,f_j,P_1,P_2)=\left\{
\begin{array}{lll}
1  & \quad \mbox{if } & (f_i,f_j) \in P_1 \cap P_2 \mbox{ or } (f_j,f_i) \in P_1 \cap P_2\\
1  & \quad \mbox{if } & (f_i,f_j) \not\in P_1 \cup P_2 \mbox{ and } (f_j,f_i) \not\in P_1 \cup P_2\\
0.5  & \quad \mbox{if } & ((f_i,f_j) \in P_1 \Delta P_2 \mbox{ and } (f_j,f_i) \not\in P_1 \cup P_2) \mbox{  or } \\
  & & ((f_j,f_i) \in P_1 \Delta P_2 \mbox{ and } (f_i,f_j) \not\in P_1 \cup P_2) \\
0  & \quad \mbox{if } & (f_i,f_j) \in P_1 \cup P_2 \mbox{ and } (f_j,f_i) \in P_1 \cup P_2\,.\\
\end{array}
\right.
\]
Here, $\Delta$ is used to denote the symmetric difference (i.e., $A \Delta B=A\cup B - A \cap B$).
In the definition of $sim(f_i,f_j,P_1,P_2)$,
\begin{itemize}
\item the first condition refers to the case where $P_1$ and $P_2$ are expressing the same order between $f_i$ and $f_j$,
\item the second condition refers to the case where both $P_1$ and $P_2$ are not expressing any order between $f_i$ and $f_j$,
\item the third condition refers to the case where one of $P_1$ and $P_2$ is expressing an order between $f_i$ and $f_j$ and the other is not expressing any order,
\item the last condition refers to the case where $P_1$ and $P_2$ are expressing opposite orders between $f_i$ and $f_j$.
\end{itemize}

Clearly, $sim(P_1,P_2)$ is $1$ when $P_1$ and $P_2$ agree on everything, and $0$ when $P_1$ and $P_2$ agree on nothing.
Finally, we define a relevance measure by $\rho((E,{\succ}_P,I),{\succ}_{P'}) = sim(\succ_P,\succ_{P'})$ for every report
$(E,\succ_P,I)\in\textit{Reports}$ and SPO $\succ_{P'}\in\textit{SPOs}(\F)$.
\smallbsq
\end{example}

%% file: queryans.tex
\section{Query Answering based on Subjective Reports}
\label{sec:QA-basic}

\begin{figure}[t]
\fbox{
\parbox{0.96\columnwidth}{
{\bf Algorithm} \reprankbasic$(\KB, Q(\vett{X}), \F, \PU, \tau, \rho, \textit{Reports}, k)$ \\
\textbf{Input: } \dpm\ ontology $\KB$, atomic query $Q(\vett{X})$, set of features $\F=\{f_1,...,f_n\}$, \\
\phantom{\textbf{Input: }} user preferences\ $\PU$, trust measure $\tau$, relevance measure $\rho$, set of reports $\textit{Reports}$, \\
\phantom{\textbf{Input: }} $k \geq 1$.\\
\textbf{Output: } Top-$k$ answers to $Q$. \\[1ex] 
\phantom{1}1.\ $\textit{RankedAns}:=\emptyset$ \\
\phantom{1}2.\ for each atom $a$ in $\textit{ans}(Q(\vett{X}),\KB)$ do begin \\
\phantom{1}3.\ \hspace*{3mm} $\textit{score}:=0$; \\
\phantom{1}4.\ \hspace*{3mm} for each report $r=(E,\succ_P,I)$ in $\textit{Reports}(\atom{a})$  do begin \\
\phantom{1}5.\ \hspace*{6mm} $\textit{trustMeasures}$:= $\tau(r)$; \\
\phantom{1}6.\ \hspace*{6mm} $\textit{score}:= \textit{score} + \rho(r,\PU) * \frac{1}{n}*
\sum_{i=1}^n E[i]* \textit{trustMeasures}[i]*\frac{1}{\textit{rank}(f_i,\PU)}$;\\
\phantom{1}7.\ \hspace*{3mm} end; \\
\phantom{1}8.\ \hspace*{3mm} $\textit{score}:=\textit{score}/|\textit{Reports}(\atom{a})|$; \\
\phantom{1}9.\ \hspace*{3mm} $\textit{RankedAns}:= \textit{RankedAns} \cup\{\<\atom{a},\textit{score}\>\}$; \\
10.\ end;\\
11.\ return top-$k$ atoms in $\textit{RankedAns}$.
}
}
\caption{
A first algorithm for computing the top-k answers to an atomic query $Q$ according to a given set of user preferences and reports on answers to $Q$.
}
\label{alg:repRankBasic}
\end{figure}

To produce a ranking based on the basic components presented in Section~\ref{sec:reports}, we must first
develop a way to combine them in a principled manner.
More specifically, the problem that we address is the following.
The user is given a \dpm\ ontology $\KB$ and has an atomic query $Q(\vett{X})$ of interest.
The user also supplies an SPO $\PU$ over the set of features $\F$.
The answers to an atomic query $Q(\vett{X})=p(\vett{X})$ over $\KB$ in {\em atom form} are defined as
$\{p(t) \mid t \in \textit{ans}(Q(\vett{X}),\KB)\}$; we still use $\textit{ans}(Q(\vett{X}),\KB)$ to denote
the set of answers in atom form.
Recall that in our setting, each ground atom $\atom{b}$ such that $\KB \models \atom{b}$ is associated with a (possibly empty) set of reports.
As we consider atomic queries, then each ground atom $\atom{a}\in \textit{ans}(Q(\vett{X}),\KB)$ is an atom entailed by $\KB$ and thus it is associated with a set of reports $\textit{Reports}(\atom{a})$.
Furthermore, each report $r\in \textit{Reports}(\atom{a})$ is associated with a trust score $\tau(r)$.
We want to rank the ground atoms in $\textit{Ans}(Q(\vett{X}),\KB)$; that is, we want to obtain a set
$\{\<\atom{a_i},\textit{score}_i\> \mid \atom{a_i} \in \textit{ans}(Q(\vett{X}),\KB)\}$ where $\textit{score}_i$
for ground atom
$\atom{a_i}$ takes into account:
\begin{itemize}
\item the set of reports $\textit{Reports}(\atom{a_i})$ associated with $\atom{a_i}$;
\item the trust score $\tau(r)$ associated with each report $r\in \textit{Reports}(\atom{a_i})$; and
\item the SPO $\PU$ over $\F$ provided by the user issuing the query.
\end{itemize}


\subsection{A Basic Approach}

A first approach to solving this problem is Algorithm \reprankbasic\ in Fig.~\ref{alg:repRankBasic}.
A~score for each atom is computed as the average of the scores of the reports associated with the atom,
where the score of a report $r=(E,\succ_P,I)$ is computed as follows:
\begin{itemize}
\item we first compute the average of the scores $E[i]$ weighted by the trust value for $E[i]$ and a value
measuring how important feature $f_i$ is for the user issuing the query (this value is given by $\textit{rank}(f_i,\PU)$);

\item then, we multiply the value computed in the previous step by $\rho(r,\PU)$, which gives a measure of how
relevant $r$ is w.r.t. $\PU$.
\end{itemize}
The following is an example of how Algorithm~\reprankbasic\ works.

\begin{figure}[t]
\centering
\begin{tabular}{|l|c|c|c|c|l|}\hline
\multicolumn{6}{|c|}{\parbox[c]{11cm}{{\bf Reports} $r_1 = (E_1, \succ_{P_1}, I_1)$ and $r_4 =  (E_4, \succ_{P_1}, I_1)$}}  \\
\multicolumn{6}{|c|}{\parbox[c]{11cm}{Relevance scores: $\rho(r_1,\PU) = \rho(r_4,\PU) = 0.125$}} \\ \hline
{\em Features}   & \; $E_1$ \; & \; $E_4$ \; & $\tau(r_1) = \tau(r_4)$ & $\succ_{P_1}$ &  $I_1$ \\ \hline
{\em loc}  & 1     & 0.9   & 1                       & \multirow{5}{*}{\includegraphics[width=2.2cm]{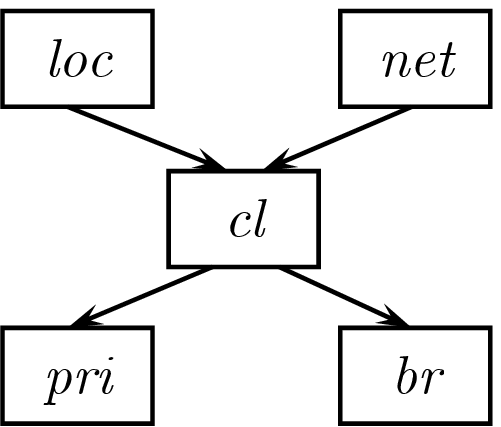}} &  \\ \cline{1-4}
{\em cl}   & 0     & 1   & 0.5                     &    & {\em Age} = 34 \\ \cline{1-4}
{\em pri}  & 0.4   & 1   & 0.25                    &    & {\em Nationality} = {\em Italian} \\ \cline{1-4}
{\em br}   & 0.1   & 1   & 0.25                    &    & {\em Type} = {\em Business} \\ \cline{1-4}
{\em net}  & 1     & 1   & 1                       &    & \\ \hline
\end{tabular}

\medskip

\begin{tabular}{|l|c|c|c|c|l|}\hline
\multicolumn{6}{|c|}{\parbox[c]{11cm}{{\bf Reports} $r_2 = (E_2, \succ_{P_2}, I_2)$ and $r_5 =  (E_5, \succ_{P_2}, I_2)$}} \\
\multicolumn{6}{|c|}{\parbox[c]{11cm}{Relevance scores: $\rho(r_2,\PU) = \rho(r_5,\PU) = 0.5$}} \\ \hline
{\em Features}   & \; $E_2$ \; & \; $E_5$ \; & $\tau(r_2) = \tau(r_5)$ & $\succ_{P_2}$ &  $I_2$ \\ \hline
{\em loc}  & 0.9   & 0.8   & 1                       & \multirow{5}{*}{\includegraphics[width=2.2cm]{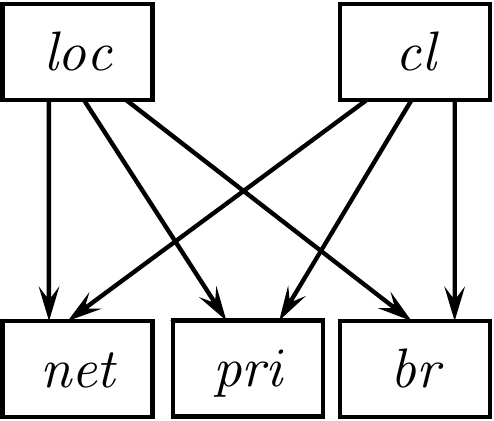}} &  \\ \cline{1-4}
{\em cl}   & 0.3   & 0.1   & 1                       &    & {\em Age} = 45 \\ \cline{1-4}
{\em pri}  & 0.2   & 0.1   & 0.5                     &    & {\em Nationality} = {\em Italian} \\ \cline{1-4}
{\em br}   & 0.5   & 0.4   & 0.5                     &    & {\em Type} = {\em Leisure} \\ \cline{1-4}
{\em net}  & 0     & 1     & 0.5                     &    & \\ \hline
\end{tabular}

\medskip

\begin{tabular}{|l|c|c|c|c|l|}\hline
\multicolumn{6}{|c|}{\parbox[c]{11cm}{{\bf Reports} $r_3 = (E_3, \succ_{P_3}, I_3)$ and $r_6 =  (E_6, \succ_{P_3}, I_3)$}} \\
\multicolumn{6}{|c|}{\parbox[c]{11cm}{Relevance scores: $\rho(r_3,\PU) = \rho(r_6,\PU) = 2^{-9}$}} \\ \hline
{\em Features}   & \; $E_3$ \; & \; $E_6$ \; & $\tau(r_3) = \tau(r_6)$ & $\succ_{P_3}$ &  $I_3$ \\ \hline
{\em loc}  & 0.85  & 0.3   & 0.0625                       & \multirow{5}{*}{\includegraphics[width=2.2cm]{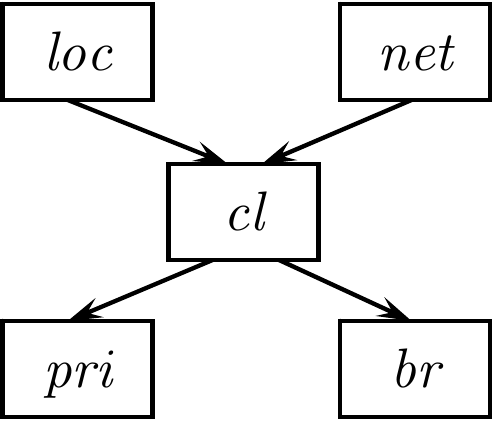}} &  \\ \cline{1-4}
{\em cl}   & 0.9   & 0.5   & 0.0313                     &    & {\em Age} = 29 \\ \cline{1-4}
{\em pri}  & 0.8   & 0.9   & 0.25                         &    & {\em Nationality} = {\em Spanish} \\ \cline{1-4}
{\em br}   & 0.8   & 0.9   & 0.25                         &    & {\em Type} = {\em Leisure} \\ \cline{1-4}
{\em net}  & 1     & 0.2   & 0.125                        &    & \\ \hline
\end{tabular}
\caption{Reports used in Examples~\ref{ex:reprankbasic} and~\ref{ex:summarize}. We assume that each pair
of reports ($r_1$--$r_4$, $r_2$--$r_5$, and $r_3$--$r_6$) was generated by the same reviewer---they thus share
the SPO and information registers.}
\label{fig:fullReports}
\end{figure}

\begin{example}
\label{ex:reprankbasic}
Consider again the setup from the running example, where we have the \dpm\ ontology from Example~\ref{ex:dpm},
the set $\textit{Reports}$ of the reports depicted in Fig.~\ref{fig:fullReports}, the SPO $\PU$ from
Fig.~\ref{fig:reports} (right), the trust measure $\tau$ defined in Example~\ref{ex:trust}, and the relevance
measure $\rho$ introduced in Example~\ref{ex:relevance}. Finally, let $Q(X) = \textit{hotel}(X)$.

Algorithm~\reprankbasic\ iterates through the set of answers (in atom form) to the query, which in this case
consists of $\{\textit{hotel}(h_1),\textit{hotel}(h_2)\}$.
For atom $\textit{hotel}(h_1)$, the algorithm iterates through the set of corresponding reports, which is
$\textit{Reports}(\textit{hotel}(h_1)) = \{r_1,r_2,r_3\}$, and maintains the accumulated score after processing
each report. For $r_1$, the score is computed as (cf.\ line~6):
\[
0.125 * \frac{1}{5} * \left( \frac{1 * 1}{1} + \frac{0 * 0.5}{1} +
\frac{0.4 * 0.25}{2} + \frac{0.1 * 0.25}{3} + \frac{1 * 1}{2} \right) = 0.03895\,.
\]
The score for $\textit{hotel}(h_1)$ after processing the three reports is approximately $0.05746$.
Analogously, assuming $\textit{Reports}(\textit{hotel}(h_2)) \eqs \{r_4,r_5,r_6\}$, the score for $\textit{hotel}(h_2)$ is approximately $0.0589$.
Therefore, the top-2 answer to $Q$ is $\langle \textit{hotel}(h_2), \textit{hotel}(h_1) \rangle$.
\smallbsq
\end{example}

The following result states the time complexity of Algorithm~\reprankbasic.
As long as both query answering and the computation of the trust and relevance measures can be done in polynomial time, \reprankbasic\ can also be done in polynomial time.

\begin{proposition}\label{pro:rankbasic-complexity}
The worst-case time complexity of Algorithm~\reprankbasic\ is \linebreak
$O(m *log\, m + (n+ |\PU|) + m *\textit{Reports}_{max}*(f_\tau+f_\rho+n)+f_{\textit{ans}(Q(\vett{X}),\KB)})$,
where
$m=|\textit{ans}(Q(\vett{X}),\KB)|$,
$\textit{Reports}_{max}=\max\{|\textit{Reports}(\atom{a})| : \atom{a} \in \textit{ans}(Q(\vett{X}),\KB)\}$,
$f_\tau$ (resp. $f_\rho$) is the worst-case time complexity of $\tau$ (resp. $\rho$), and
$f_{\textit{ans}(Q(\vett{X}),\KB)}$ is the data complexity of computing $\textit{ans}(Q(\vett{X}),\KB)$.
\end{proposition}

In the next section, we explore an alternative approach to applying the trust and relevance measures to top-k query answering.

\subsection{A Different Approach to using Trust and Relevance}

A more complex approach consists of using the trust and relevance scores provided by the respective measures
in a more fine-grained manner. One way of doing this is via the following steps (more details on each of them
are given shortly):
\begin{enumerate}
\item Keep only those reports that are most relevant to the user issuing the query, that is, those reports that are relevant enough to $\PU$ according to a relevance measure $\rho$;

\item consider the most relevant reports obtained in the previous step and
use the trust measure given by the user to produce scores adjusted by the trust measure; and

\item for each atom, compute a single score by combining the scores computed in the previous step with $\PU$.
\end{enumerate}

The first step can simply be carried out by checking, for each report $r$, if $\rho(r,\PU)$ is above a certain
given threshold.
One way of doing the second step is described in Algorithm~\summarizeReports\ (Fig.~\ref{alg:trustAdjScore}),
which takes a trust measure $\tau$, a set of reports $\textit{Reports}$ (for a certain atom), and a function $\textit{collFunc}$.
The algorithm processes each report in the input sets by building a histogram of average (trust-adjusted) reported
values for each of the $n$ features with ten possible ``buckets'' (of course, this can be easily generalized to any
number of buckets); for each report, the algorithm applies the trust measure  to update each feature's histogram. Once all of the reports
are processed, the last step is to collapse the histograms into a single value---this is done by applying the
$\textit{collFunc}$ function, which could simply be defined as the computation of a weighted average for each feature.
This single value is finally used to produce the output, which is a tuple of $n$ scores.
The following example illustrates how \summarizeReports\ works.

\begin{example}
\label{ex:summarize}
Let us adopt again the setup from Example~\ref{ex:reprankbasic}.
Suppose we want to keep only those reports for which the relevance score is above $0.1$ (as per the first step of our more complex approach).
Recall that the set of answers to $Q$ is $\{\textit{hotel}(h_1),\textit{hotel}(h_2)\}$ and there are six associated reports.
Among them, we keep only reports $r_1$, $r_2$, $r_4$, and $r_5$.
Algorithm~\summarizeReports\ will have $\textit{Reports} = \{r_1,r_2\}$ when called for $\textit{hotel}(h_1)$. The histograms built during this
call are as follows:
\begin{itemize}
\item $\loc$: value 0.95 in bucket $[0.9,1]$;

\item $\cl$: value 1 in bucket $[0.5,0.6]$ and value 0.3 in bucket $[0.9,1]$;

\item $\pri$: value 0.8 in bucket $[0.2,0.3)$ and value 0.2 in bucket $[0.5,0.6)$;

\item $\br$: value 0.1 in bucket $[0.2,0.3)$ and value 0.5 in bucket $[0.5,0.6)$; and

\item $\net$: value 0.6 in bucket $[0.6,0.7)$ and value 1 in bucket $[0.9,1]$.
\end{itemize}
Assuming that function $\textit{collFunc}$ disregards the values in the bucket corresponding to the lowest trust
value (if more than one bucket is non-empty), and takes the average of the rest, we have the following result
tuple as the output of \summarizeReports:
$(0.95, 0.3, 0.2, 0.5, 1)$. Analogously, we have tuple $(0.85, 0.1, 0.1, 0.4, 1)$ for tuple
$\textit{hotel}(h_2)$ after
calling \summarizeReports\ with $\textit{Reports} = \{r_4,r_5\}$.
\smallbsq
\end{example}

The following proposition states the time complexity of Algorithm~\summarizeReports.
As long as the trust measure and the $\textit{collFunc}$ function can be computed in polynomial time, Algorithm~\summarizeReports\ is polynomial time too.

\begin{proposition}\label{pro:summarize-complexity}
The worst-case time complexity of Algorithm~\summarizeReports\ is
$O(|\textit{Reports}|*(f_\tau+n)+n*f_{\textit{collFunc}})$,
where $f_\tau$ (resp. $f_{\textit{collFunc}}$) is the worst-case time complexity of $\tau$ (resp. $\textit{collFunc}$).
\end{proposition}

The following example explores a few different ways in which function $\textit{collFunc}$ used in
Algorithm~\summarizeReports\ might be defined.

\begin{example}
\label{ex:collFunc}
One way of computing $\textit{collFunc}$ is shown in Example~\ref{ex:summarize}.
There can be other reasonable ways of collapsing the histogram for a feature into a single value.
E.g., $\textit{collFunc}$ might compute the average across all buckets ignoring the trust
measure so that no distinction is made among buckets, i.e.,
$\textit{collFunc}(\textit{hists}[i])= \frac{\sum_{b=1}^{10} \textit{hists}[i](b)}{10}$.
Alternatively, the trust measure might be taken into account by giving a  weight $w_b$ to each bucket
$b$ (e.g., the weights might be set in such a way that buckets corresponding to higher trust scores
have a higher weight, that is, $weight_i < weight_j$ for $i < j$).
In this case, the histogram might be collapsed as follows $\textit{collFunc}(\textit{hists}[i])=
\frac{\sum_{b=1}^{10} w_b * \textit{hists}[i](b)}{10}$.
We may also want to apply the above strategies but ignoring the first $k$ buckets (for which
the trust score is lower).
Function $\textit{collFunc}$ can also be extended so that the number of elements associated with a
bucket is taken into account.
\smallbsq
\end{example}

Thus, the second step discussed above gives $n$ scores (adjusted by the trust measure) for each ground atom.
Recall that the third (and last) step of the approach adopted in this section is  to compute a score for each atom by combining the scores computed in the previous step with $\PU$.
One simple way of doing this is to compute the weighted average of such scores where the weight of the $i$-th score is the inverse of the rank of feature $f_i$ in $\PU$.

\begin{figure}[t]
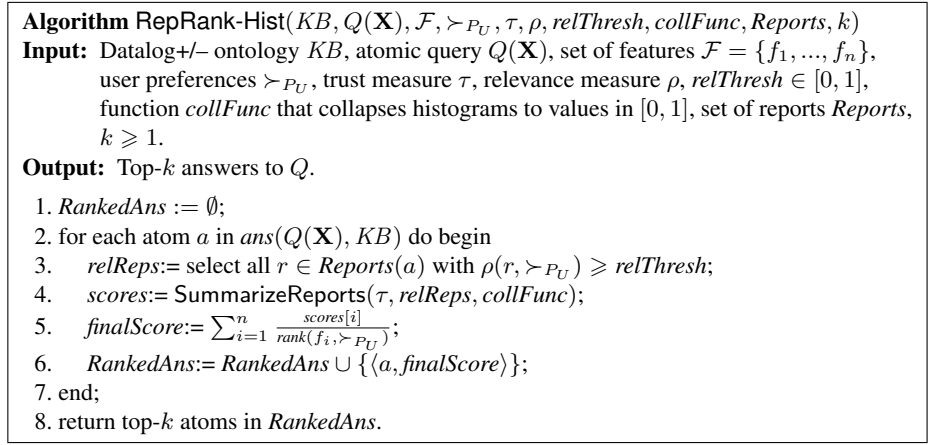

\fbox{
\parbox{0.96\columnwidth}{
{\bf Algorithm} \reprankhist$(\KB, Q(\vett{X}), \F, \PU, \tau, \rho, \textit{relThresh}, \textit{collFunc}, \textit{Reports}, k)$ \\
\textbf{Input: } \dpm\ ontology $\KB$, atomic query $Q(\vett{X})$, set of features $\F=\{f_1,...,f_n\}$, \\
\phantom{\textbf{Input: }} user preferences\ $\PU$, trust measure $\tau$, relevance measure $\rho$, $\textit{relThresh} \in [0,1]$,  \\
\phantom{\textbf{Input: }}  function $\textit{collFunc}$ that collapses histograms to values in $[0,1]$, set of reports $\textit{Reports}$, \\
\phantom{\textbf{Input: }} $k \geq 1$. \\
\textbf{Output: } Top-$k$ answers to $Q$. \\[1ex] 
\phantom{1}1.\ $\textit{RankedAns}:=\emptyset$; \\
\phantom{1}2.\ for each atom $a$ in $\textit{ans}(Q(\vett{X}),\KB)$ do begin \\
\phantom{1}3.\ \hspace*{3mm} $\textit{relReps}$:= select all $r \in \textit{Reports}(\atom{a})$ with $\rho(r,\PU) \geq \textit{relThresh}$; \\
\phantom{1}4.\ \hspace*{3mm} $\textit{scores}$:= $\summarizeReports(\tau, \textit{relReps}, \textit{collFunc})$; \\
\phantom{1}5.\ \hspace*{3mm} $\textit{finalScore}$:= $\sum_{i=1}^n\frac{\textit{scores}[i] }{\textit{rank}(f_i,\PU)}$; \\
\phantom{1}6.\ \hspace*{3mm} $\textit{RankedAns}$:= $\textit{RankedAns} \cup \{\<\atom{a},\textit{finalScore}\>\}$; \\
\phantom{1}7.\ end; \\
\phantom{1}8.\ return top-$k$ atoms in $\textit{RankedAns}$.
}
}
\caption{
Algorithm for computing the top-k answers to an atomic query $Q$ according to a given set of user preferences and reports on answers to $Q$.
}
\label{alg:repRankHist}
\end{figure}

Algorithm~\reprankhist\ (Figure~\ref{alg:repRankHist}) is the complete algorithm that combines the three steps discussed thus far. The following continues the running example to show the result of applying this algorithm.

\begin{figure}[t]
\fbox{
\parbox{0.96\columnwidth}{
{\bf Algorithm} \summarizeReports$(\tau, \textit{Reports}, \textit{collFunc})$ \\
\textbf{Input: } Trust measure $\tau$, set of reports $\textit{Reports}$, and
function $\textit{collFunc}$ that collapses \\
\phantom{\textbf{Input: }} histograms to values in $[0,1]$.\\
\textbf{Output: } Scores representing the trust-adjusted average from $\textit{Reports}$. \\[1ex]
\phantom{1}1.\ Init.\ $\mathit{hists}$ as an $n$-array of empty mappings with keys:
$\{[0,0.1),[0.1,0.2), ..., [0.9,1]\}$\\
\phantom{11.}\ and values of type $[0,1]$, where $n = |\F|$ (we use values $1, ..., 10$ to denote the keys). \\
\phantom{1}2.\ Initialize array $\textit{bucketCounts}$ of size $n \times 10$ with value $0$ in all positions; \\
\phantom{1}3.\ for each report $r = (E,\succ_P,I) \in \textit{Reports}$ do begin; \\
\phantom{1}4.\ \hspace*{3mm} $\textit{trustMeasures}$:= $\tau(r)$; \\
\phantom{1}5.\ \hspace*{3mm} for $i = 1$ to $n$ do begin \\
\phantom{1}6.\ \hspace*{6mm} let $b$ be the key for $\mathit{hists}[i]$ under which $\textit{trustMeasures}[i]$ falls;  \\
\phantom{1}7.\ \hspace*{6mm} $\textit{hists}[i](b)$:=
$\frac{\mathit{hist}[i](b) * \textit{bucketCounts}[i][b] + E[i]}{\textit{bucketCounts}[i][b] + 1}$; \\
\phantom{1}8.\ \hspace*{6mm} $\textit{bucketCounts}[i][b]$++; \\
\phantom{1}9.\ \hspace*{3mm} end; \\
10.\ end;  \\
11.\ Initialize $\textit{Res}$ as an $n$-array; \\
12.\ for $i = 1$ to $n$ do begin \\
13.\ \hspace*{3mm} $\textit{Res}[i]$:= $\textit{collFunc}(\textit{hists}[i])$; \\
14.\ end; \\
15.\ return $\mathit{Res}$.
}
}
\caption{
This algorithm takes a set of reports for a single entity and computes an $n$-array of scores obtained by combining the reports with their trust measure.
}
\label{alg:trustAdjScore}
\end{figure}

\begin{example}
\label{ex:reprank-hist}
Let us adopt once again the setup from Example~\ref{ex:reprankbasic}, but this time applying Algorithm~\reprankhist.
Suppose $\textit{collFunc}$ is the one discussed in Example~\ref{ex:summarize}  and thus Algorithm~\summarizeReports\ returns the scores
$(0.95, 0.3, 0.2, 0.5, 1)$ for $\textit{hotel}(h_1)$ and the scores $(0.85, 0.1, 0.1, 0.4, 1)$ for $\textit{hotel}(h_2)$.
Algorithm~\reprankhist\ computes a score for each atom by performing a weighted average of the scores in these tuples, which
results in:
\[
\langle \textit{hotel}(h_1), 2.0166 \rangle \; \text{and} \;
\langle \textit{hotel}(h_2), 1.6333 \rangle.
\]
Therefore, the top-2 answer to query $Q$ is
$\langle \textit{hotel}(h_1), \textit{hotel}(h_2) \rangle$.
\smallbsq
\end{example}

Note that the results from Examples~\ref{ex:reprankbasic} and~\ref{ex:reprank-hist} differ in the way they
order the two tuples; this is due to the way in which relevance and trust scores are used in each algorithm---the more
fine-grained approach adopted by Algorithm~\reprankhist\ allows it to selectively use both kinds of values to generate
a more informed result.

\begin{proposition}
\label{pro:rankhistc-complexity}
The worst-case time complexity of Algorithm~\reprankhist\ is: \linebreak
$
O(m *log\, m + (n+ |\PU|) + m * (\textit{Reports}_{max}*f_\rho + f_{\textit{sum}} + n)+f_{\textit{ans}(Q(\vett{X}),\KB)})
$, where
$m=|\textit{ans}(Q(\vett{X}),\KB)|$,
$\textit{Reports}_{max}=\max\{|\textit{Reports}(\atom{a})| : \atom{a} \in \textit{ans}(Q(\vett{X}),\KB)\}$,
$f_\rho$ is the worst-case time complexity of $\rho$,
$f_{\textit{sum}}$ is the worst-case time complexity of Algorithm~\summarizeReports\
as per Proposition~\ref{pro:summarize-complexity}, and
$f_{\textit{ans}(Q(\vett{X}),\KB)}$ is the data complexity of computing $\textit{ans}(Q(\vett{X}),\KB)$.
\end{proposition}
As a corollary to Propositions~\ref{pro:rankbasic-complexity} and~\ref{pro:rankhistc-complexity}, we have the
following result.

\begin{theorem}
\label{theo:poly-basic}
If the input ontology belongs to the guarded fragment of \dpm, then
Algorithms~\reprankbasic\ and~\reprankhist\ run
in polynomial time in the data complexity.
\end{theorem}

Thus far, we have considered atomic queries.
As each ground atom $a$ such that $\KB\models a$ is associated with a set of reports and every ground atom $b$ in $\textit{ans}(Q(\vett{X}),\KB)$ is such that $\KB\models b$, then reports can be associated with query answers in a natural way.
We now introduce a class of queries more general than the class of atomic queries for which the same property holds.
A \emph{\sq\ query} is a conjunctive query $Q(\vett{X})=\exists\vett{Y}\,\Phi(\vett{X},\vett{Y})$ where $\Phi(\vett{X},\vett{Y})$ contains exactly one atom of the form $p(\vett{X})$, called \emph{distinguished} atom (i.e., an atom whose variables are
the query's free variables).
For instance, $Q(X) = \textit{hotel}(X) \wedge \textit{locatedIn}(X,\textit{oxford})$ is a \sq\ query where
$\textit{hotel}(X)$ is the distinguished atom.
The answers to a \sq\ query $Q(\vett{X})$ over $\KB$ in {\em atom form} are defined as $\{p(t) \mid t \in \textit{ans}(Q(\vett{X}),\KB)\}$ where the distinguished atom is of the form $p(\vett{X})$; we still use $\textit{ans}(Q(\vett{X}),\KB)$ to denote the set of answers in atom form.
Clearly, for each atom $a$ in $\textit{ans}(Q(\vett{X}),\KB)$, it is the case that $\KB\models a$.

%% file: gen-reports.tex
\section{Towards more General Reports}
\label{sec:general}

In the previous section we considered the setting where reports are associated with ground atoms $\atom{a}$ such that $\KB\models\atom{a}$. This setup is limited, since it does not allow to express the fact that certain reports
may apply to whole {\em sets} of atoms---this is necessary to model certain kinds of opinions often found in reviews,
such as ``accommodations in Oxford are expensive''.
We now generalize the framework presented in Sections~\ref{sec:reports}
and~\ref{sec:QA-basic} to contemplate this kind of reports.

\begin{definition}{\rm
\label{def:gen-report}
A {\em  generalized report} (g-report, for short) is a pair $gr = (r, Q(\vett{X}))$,  where $r$ is a report and $Q(\vett{X})$ is a \sq\ query, called the {\em descriptor} of $gr$.
}\end{definition}
We denote with $\textit{g-Reports}$ the universe of g-reports.
Intuitively, given an ontology $\KB$, a g-report $(r, Q(\vett{X}))$ is used to associate report $r$ with
every atom $a$ in $\textit{ans}(Q(\vett{X}),\KB)$---recall that $\KB\models a$ and thus general reports
allow us to assign a report to a set of atoms entailed by $\KB$.

Clearly, a report for a ground atom $\atom{a}$ as defined in Definition~\ref{def:report} is a special case of a g-report in which the only answer to the descriptor is $\atom{a}$.

\begin{example}
\label{ex:gen-report}
Consider our running example from the accommodations domain and suppose we want to associate a certain report $r$ with all accommodations in the city of Oxford.
This can be expressed with a g-report $(r,Q(X))$ where
$Q(X) = \textit{accom}(X) \wedge \textit{locatedIn}(X,\textit{oxford})$ with
descriptor $\textit{accom}(X)$.
\end{example}

Intuitively, a g-report $gr = (r,Q(\vett{X}))$ is a report associated with a set of atoms, i.e., the set of atoms
in $\textit{ans}(Q(\vett{X}), \KB))$. A simple way of handling this generalization would be to associate
report $r$ with every atom in this set.
Note that, as in the non-generalized case, it might be the case that two or more g-reports assign two distinct
reports to the same ground atom.
E.g.,
we may have a g-report $(r,Q(X))$, where
$Q(X) = \textit{accom}(X) \wedge \textit{locatedIn}(X,\textit{oxford})$, expressing that $r$ applies to
all accommodations in Oxford, and another g-report $(r',Q'(X))$, where
$Q'(X) = \textit{accom}(X) \wedge \textit{hotel}(X)$, expressing that $r'$ applies to all accommodations that are hotels.
In our running example, we would simply associate both $r$ and $r'$ to
$\textit{accom}(h_i)$, $\textit{accom}(h_2)$, and $\textit{accom}(a_2)$.

In the approach just described, the reports coming from different g-reports are treated in the same way---they
all have the same impact on the common atoms.
Another possibility is to determine when a g-report is in some sense {\em more specific} than another and take
such a relationship into account (e.g., more specific g-reports should a greater impact when computing the ranking
over atoms).
We consider this kind of scenario in the following section.


\subsection*{Leveraging the Structural Properties of Ontologies}

We now study two kinds of structure that can be leveraged from knowledge contained in the ontology.
The first is based on the notion of {\em hierarchies}, which are useful in capturing the influence of
reports in ``is-a'' type relationships. As an example, given a query requesting a ranking over hotels
in Oxfordshire,
a report for all hotels in Oxford should have a higher
impact on the calculation of the ranking than a report for all {\em accommodations} in the UK---in particular, the
latter might be ignored altogether since it is too general. The second kind of structure is based on
identifying subset relationships between the atoms associated with the descriptors in g-reports. For instance,
a report for all hotels in Oxford is more general than a report for all hotels in Oxford city center, since
the former is a superset of the latter.

In the following, we define a partial order among reports based on these notions. We begin by defining
hierarchical TGDs.

\begin{definition}{\rm
\label{def:hierarchicalOnt}
A set of linear TGDs $\Sigma_T$ is said to be {\em hierarchical} iff for every
$p(\vett{X}) \rightarrow \exists \vett{Y} q(\vett{X},\vett{Y}) \in \Sigma_T$ we have that
$\textit{features}(p) \subseteq \textit{features}(q)$ and there does not exist
database $D$ over $\R$ and TGD in $\Sigma_T$ of the form
$p'(\vett{X}) \rightarrow \exists \vett{Y} r(\vett{X},\vett{Y})$ such that
$p(\vett{X})$ and $p'(\vett{X})$ share ground instances relative to $D$.
}\end{definition}

In the rest of this section, we assume that all ontologies contain a (possibly empty) subset of hierarchical TGDs.
Furthermore, given ontology $\KB = (D, \Sigma)$ where $\Sigma_H \subseteq \Sigma$ is a set of hierarchical TGDs,
and
two ground atoms $a, b$, we say that $a$ {\em is-a} $b$ iff $\textit{chase}(\{a\}, \Sigma_H) \models b$.
For instance, in Example~\ref{ex:dpm}, set $\{\sigma_1, \sigma_2, \sigma_3, \sigma_4\} \subseteq \Sigma$ is
a hierarchical set of TGDs (assuming that the conditions over the features hold).

Given tuples of features
$\F$ and $\F'$ such that $\F \subseteq \F'$ and vectors $E$ and $E'$ over
the domains of $\F$ and $\F'$, respectively, we say that $E'$ is a particularization of $E$,
denoted $E' = \textit{part}(E)$ iff $E'[f] = E[f]$ if $f \in F \cap \F'$ and $E'[f] = -$ otherwise.

\begin{definition}{\rm
\label{def:spec}
Let $\KB = (D,\Sigma)$ be a \dpm\ ontology, $a$ be a ground atom such that $\KB \models a$,
and $gr = (r, Q(\vett{X}))$ be a g-report with $r = (E,\succ_P,I)$.
If there exists a ground atom $b \in \textit{Ans}(Q(\vett{X}),\KB)$ such that $a$ {\em is-a} $b$
then we say that g-report $gr' = ((E',\succ_P,I), a)$, with $E' = \textit{part}(E)$,
is a {\em specialization} of $gr$ for $a$.
}\end{definition}

Clearly, a g-report is always a specialization of itself for every atom in the answers to
its descriptor.

\begin{example}
\label{ex:specialization}
Let $\F_1$ be the set of features for predicate $\textit{hotel}$ presented in Example~\ref{ex:reports},
and let $\F_2 = \langle \loc, \cl, \pri, \br, \net, \kf \rangle$ be the set of features for predicate
$\textit{apthotel}$, where $\kf$ denotes ``kitchen facilities''.

Let $gr \eqs (r_1, Q(X))$ be a g-report, where $r_1$ is the report from Figure~\ref{fig:fullReports}
and $Q(X) = \textit{hotel}(X) \wedge \textit{locatedIn}(X,\textit{oxford})$.
If we consider
$a = \textit{apthotel}(a_2)$ and
$b = \textit{hotel}(a_2)$, clearly we have that
$b \in \textit{Ans}(Q(X),\KB)$ and $a$ is-a $b$.
Therefore, a specialization of $gr$ for $a$ is $gr' = ((E',\succ_{P_1},I_1), a)$, where
$E' = \langle 1, 0, 0.4, 0.1, 1, - \rangle$.\smallbsq
\end{example}

\begin{figure}[t]
\centering
\fbox{
\parbox{0.7\columnwidth}{
\begin{tabular}{lll}
$gr_1$ & = & $\big((r_1,\succ_{P_1}, I_1), \underline{\textit{hotel}}(X) \wedge \textit{locatedIn}(X, \textit{oxford})\big)$ \\
$gr_2$ & = & $\big((r_2,\succ_{P_2}, I_2), \underline{\textit{hotel}}(X) \wedge \textit{locatedIn}(X, \textit{cambridge})\big)$ \\
$gr_3$ & = & $\big((r_3,\succ_{P_3}, I_3), \underline{\textit{apthotel}}(X) \wedge \textit{locatedIn}(X, \textit{oxfordCenter})\big)$ \\
$gr_4$ & = & $\big((r_4,\succ_{P_1}, I_1), \underline{\textit{hotel}}(X) \wedge \textit{locatedIn}(X, \textit{oxfordCenter})\big)$
\end{tabular}
}}
\caption{A set of general reports (distinguished atoms in the descriptors are underlined).}
\label{fig:g-reports}
\end{figure}

\begin{definition}{\rm
\label{def:g-rep:more-gen}
Given g-reports $gr_1=(r_1,Q_1(\vett{X_1}))$ and $gr_2=(r_2,Q_2(\vett{X_2}))$, we say that $gr_1$ is \emph{more general than}
$gr_2$, denoted $gr_2 \mg gr_1$, iff either
(i) $\textit{Ans}(Q_2(\vett{X_2}),\KB) \subseteq \textit{Ans}(Q_1(\vett{X_1}),\KB)$; or
(ii) for each $a \in \textit{Ans}(Q_2(\vett{X_2}),\KB)$ there exists $b \in \textit{Ans}(Q_1(\vett{X_1}),$ $\KB)$ such that
$a$ is-a $b$.
If $gr_1\mg gr_2$ and $gr_2\mg gr_1$, we say that $gr_1$ and $gr_2$ are \emph{equivalent}, denoted $gr_1 \equiv gr_2$.
}\end{definition}

\begin{example}
\label{ex:g-reports}
Consider the g-reports in Figure~\ref{fig:g-reports} and the database in the running example with the
addition of atoms $\textit{hotel}(h_3)$ and $\textit{locatedIn}(h_3,\textit{cambridge})$.
We then have:

\smallskip
\noindent
-- $gr_1 \mg gr_4$ since
$\{\textit{hotel}(h_2), \textit{hotel}(a_2)\} \subseteq
\{\textit{hotel}(h_1), \textit{hotel}(h_2), \textit{hotel}(a_2)\}$;

\smallskip
\noindent
-- $gr_1 \mg gr_3$ since for atom
$\textit{apthotel}(a_2)$ (the only answer for the descriptor in $gr_3$) there exists atom $\textit{hotel}(a_2)$
in the answer to descriptor in $gr_1$ and $\textit{apthotel}(a_2)$ is-a $\textit{hotel}(a_2)$; and

\smallskip
\noindent
-- $gr_4$ is incomparable to all other reports, since neither condition from Definition~\ref{def:g-rep:more-gen}
is satisfied. \smallbsq
\end{example}

The ``more general than'' relationship between g-reports is useful for defining a partial order for the set of
reports associated with a given ground atom. This partial order can be defined as follows:
$gr_1 \sim gr_2$ iff $gr_1 \equiv gr_2$ and $gr_1 \succ gr_2$ iff $gr_1 \mg gr_2$.
Here, $a \sim b$ denotes the equivalence between $a$ and $b$. 

\begin{definition}{\rm
\label{def:g-rep-weightfunc}
A {\em weighting function} for g-reports is any function
$\omega: \textit{g-Reports} \rightarrow [0,1]$ such that:
(i) if $gr_1 \succ gr_2$ then $\omega(gr_1) > \omega(gr_2)$; and
(i) if $gr_1 \sim gr_2$ then $\omega(gr_1) = \omega(gr_2)$.
}\end{definition}

For example, one possible weighting function is defined as $\omega(gr) = 2^{-\textit{rank}(gr,\succ)+1}$.

%% file: summary.tex
\section{Related Work}
\label{sec:rel}

The study of preferences has been carried out in many disciplines; in computer science, the developments 
that are most relevant to our work is in the incorporation of preferences into query
answering mechanisms. To date (and to our knowledge), the state of the art in this respect is centered 
around relational databases and, recently, in ontological languages for the Semantic Web~\cite{ijcai13}.
The seminal work in preference-based query answering was that of~\cite{lacroix1987preferences},
in which the authors extend the SQL language to incorporate user preferences. The preference formula formalism 
was introduced in~\cite{Chomicki:2003} as a way to embed a user's preferences into SQL. An important development
in this line of research is the well-known {\em skyline} operator, which was first introduced in~\cite{Borzsonyi:2001}.
A recent survey of preference-based query answering formalisms is provided in~\cite{Stefanidis:2011}.
Studies of preferences related to our approach have also been done in classical logic 
programming~\cite{govindarajan1995preference,govindarajan2001preference} as well as answer set programming 
frameworks~\cite{brewka07}.

The present work can be considered as a further development of the Pref\dpm\ framework presented in~\cite{ijcai13}, where 
we develop algorithms to answer skyline queries, and their generalization to $k$-rank queries, over classical 
\dpm\ ontologies. The main difference between Pref\dpm\ and the work presented here is that PrefData\-log+/-- assumes that
a model of the user's preferences are given at the time the query is issued. On the other hand, we make no 
such assumption here; instead, we assume that the user only provides some very basic information regarding their preferences
over certain features, and that they have access to a set of reports provided by other users in the past. In a sense,
this approach is akin to building an ad hoc model {\em on the fly} at query time and using it to provide a ranked 
list of results.
 
Finally, this work is closely connected to the study and use of provenance in information systems and, in particular, 
the Semantic Web and social media~\cite{Moreau:2010,Barbier2013}.
Provenance information describes the history of data and information in its life cycle. Research in provenance distinguishes between data and workflow provenance \cite{Buneman2013}. The former explores the data flow within (typically, database) applications  in a fine-grained way, 
while the latter is coarse-grained and does not consider the flow of data within the involved applications. In this work, we propose a new kind of provenance that is closely related to data provenance, but does not fit into the why,  how, and where provenance framework typically considered in data provenance research \cite{Cheney2009}. We take into account (in a fine-grained way) where evaluations and reports within a social media system are coming from (i.e., information about who has issued the report and what his/her preferences were) and use this information to allow users to make informed and~pro\-ve\-nance-based decisions. To our knowledge, this is the first study of a direct application of provenance of reports of this kind found in online reviews to query answering.
 
\section{Summary and Outlook}
\label{sec:conc}

In this paper, we have studied the problem of preference-based query answering in Data\-log+/-- ontologies
under the assumption that the user's preferences are informed by a set of subjective reports representing opinions
of others---such reports model the kind of information found, e.g.,  in online reviews of products,
places, and services. We have first introduced a basic approach, in which reports are assigned to ground atoms. 
We have proposed two ranking algorithms
using trust and relevance functions in order to model the different impact that reports should have on the 
user-specific ranking by taking into account the differences and similarities between the user's  
preferences over basic features and those of the person writing the report, as well as the person's self-reported
characteristics (such as age, gender, etc.). As a generalization, we have then extended 
reports to apply to entire sets of atoms so that they can model more general opinions. 
Apart from the naive approach of simply replicating the general report for each individual atom that it pertains to, we have 
proposed a way to use the information in the knowledge base to assign greater weights to more specific reports.

Much work remains to be done in this line of research, for instance, 
exploring conditions over the trust and relevance functions 
to allow pruning of reports, applying more sophisticated techniques to judging the impact of generalized reports, 
and the application of existing techniques to allow
the obtention of reports from the actual information available in reviews on the Web.
We also plan to implement our algorithms and~evaluate them over synthetic and real-world data. Finally, another topic for future research is to formally investigate the relationship between well-known data provenance frameworks and the preference-based provenance framework  presented in this paper.